\documentclass[default,iicol]{sn-jnl}

\jyear{2022}%

\theoremstyle{thmstyleone}%

\theoremstyle{thmstyletwo}%

\theoremstyle{thmstylethree}%

\usepackage{framed,multirow}
\usepackage{amssymb}
\usepackage{latexsym}
\usepackage{url}
\usepackage{xcolor}
\definecolor{newcolor}{rgb}{.8,.349,.1}
\usepackage{hyperref}
\usepackage[utf8x]{inputenc}
\usepackage{multicol}
\usepackage{caption}
\usepackage{hhline}
\usepackage{multirow}
\usepackage{xspace}
\usepackage{graphicx}
\usepackage{amsmath}
\usepackage{rotating}
\usepackage{xcolor}
\usepackage{framed}
\usepackage{subcaption}
\usepackage{balance}
\usepackage{booktabs}
\usepackage{bm}
\usepackage{bbm}
\usepackage{enumitem}
\usepackage{lipsum}
\usepackage[export]{adjustbox}
\usepackage{tabu}

\newcommand{\ie}{\textit{i}.\textit{e}.}
\newcommand{\eg}{\textit{e}.\textit{g}.,}
\newcommand{\etal}{\textit{et al}.}

\newcommand{\tit}[1]{\smallbreak\noindent\textbf{#1.}}

\begin{document}

\title[Article Title]{Boosting Modern and Historical Handwritten Text Recognition with Deformable Convolutions}

\author*{\fnm{Silvia} \sur{Cascianelli}}\email{silvia.cascianelli@unimore.it}
\author{\fnm{Marcella} \sur{Cornia}}\email{marcella.cornia@unimore.it}
\author{\fnm{Lorenzo} \sur{Baraldi}}\email{lorenzo.baraldi@unimore.it}
\author{\fnm{Rita} \sur{Cucchiara}}\email{rita.cucchiara@unimore.it}

\affil{\orgdiv{Department of Engineering ``Enzo Ferrari''}, \orgname{University of Modena and Reggio Emilia},\\\orgaddress{\street{Via Pietro Vivarelli, 10}, \city{Modena}, \postcode{41125}, \country{Italy}}}

\abstract{Handwritten Text Recognition (HTR) in free-layout pages is a challenging image understanding task that can provide a relevant boost to the digitization of handwritten documents and reuse of their content. The task becomes even more challenging when dealing with historical documents due to the variability of the writing style and degradation of the page quality. 
State-of-the-art HTR approaches typically couple recurrent structures for sequence modeling with Convolutional Neural Networks for visual feature extraction. Since convolutional kernels are defined on fixed grids and focus on all input pixels independently while moving over the input image, this strategy disregards the fact that handwritten characters can vary in shape, scale, and orientation even within the same document and that the ink pixels are more relevant than the background ones.
To cope with these specific HTR difficulties, we propose to adopt deformable convolutions, which can deform depending on the input at hand and better adapt to the geometric variations of the text.  
We design two deformable architectures and conduct extensive experiments on both modern and historical datasets. Experimental results confirm the suitability of deformable convolutions for the HTR task. }


\maketitle


\section{Introduction}
\label{sec:introduction}
Handwritten Text Recognition (HTR) aims at automatically understanding the content of a handwritten document by providing a natural language transcription of its textual content. Because of the key role it can play in the automatization of the digitization of documents, it can be applied in automated services and any modern document processing pipeline. On the other hand, HTR is also applied in the field of Digital Humanities~\cite{fontanella2020pattern} for the transcription of historical documents and to enable search and retrieval applications in ancient corpora, which would not be easily accessible otherwise~\cite{santoro2020using}.

Despite the advancements in Optical Character Recognition (OCR), HTR remains a challenging task.
The task, originally tackled via Hidden Markov Models built upon heuristic visual features~\cite{marti2000handwritten,krevat2006improving}, is currently performed with Deep Neural Network-based approaches~\cite{shi2016end,puigcerver2017multidimensional,quiros2018hmms}. 
The most common strategy entails employing a Convolutional Neural Network (ConvNet) and a Recurrent Neural Network(RNN)  to represent the input manuscript image and a Connectionist Temporal Classifier (CTC) to generate the output text sequence~\cite{puigcerver2017multidimensional}.
The visual feature extraction phase of these approaches relies on standard convolutional layers in which features from the input image are extracted by sliding kernels with a fixed regular grid and constant parameters.
As a consequence of the constancy of convolutional weights, pixels in an image neighborhood are encoded according to their relative positions rather than the content of the neighborhood itself. 
This can be sub-optimal if considering the characteristics of HTR images, which contain handwritten characters and words. In fact, in these images, only the ink pixels are relevant for recognition, while the background ones can potentially contain misleading nuisances, especially in ancient documents. Moreover, handwritten documents typically feature highly varying characters, which cannot be effectively modeled with standard fixed-shape convolution kernels without performing ad hoc preprocessing and data augmentation.

\begin{figure}[t]
\centering
\includegraphics[width=\columnwidth]{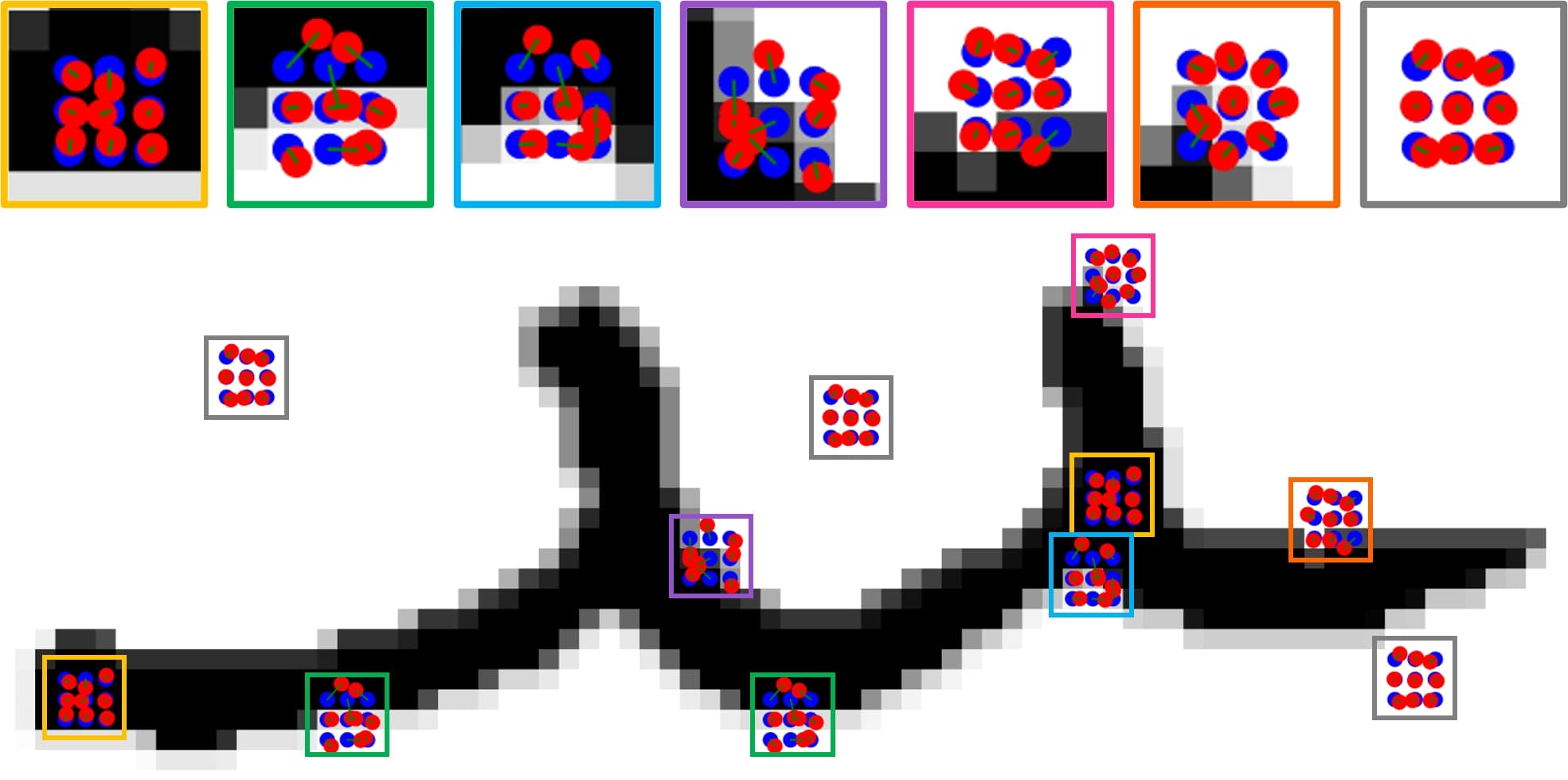}
\caption{Sampling grid of a standard convolutional kernel (in blue) and a deformable convolutional kernel (in red) when applied over a handwritten character. Deformable convolutions apply translation vectors to the kernel grid and adapt better to handwritten strokes (best seen in color).}
\label{fig:overview}
\end{figure}

Motivated by the above considerations, in this paper, we propose to employ deformable convolution operators~\cite{dai2017deformable} in HTR architectures. In deformable convolutions, the regular kernel grid is altered by adding a translation vector to the location of each kernel element. Translations vectors are computed in a content-dependent manner so that the kernel grid can be deformed depending on the input of the convolutional layer itself. 
As a consequence, deformable convolutions can adapt to the input geometric variations and part deformations, making them potentially more suitable for dealing with HTR images compared to standard convolutions. 
This kind of convolution has been originally proposed to tackle the object recognition task and, to the best of our knowledge, its usage in HTR has been explored only in our preliminary work~\cite{cojocaru2020watch}, where we claim that its kernel adaptability (see Figure~\ref{fig:overview}) can help to improve the efficiency and the performance in the task. In this work, we deepen our analysis and extend it to HTR on historical manuscripts.
To demonstrate the effectiveness of using deformable kernels, we design two different deformable architectures and conduct extensive experiments on both modern and historical datasets for handwritten text recognition. Experimental results will demonstrate that deformable convolutions are a suitable operator for HTR networks and provide consistent improvements over standard convolutions.

The rest of this paper is organized as follows. In Section~\ref{sec:related}, we provide a brief overview of the HTR literature. Then, in Section~\ref{sec:method}, we describe the proposed approach and how deformable convolutions are applied to existing HTR networks. Finally, in Section~\ref{sec:experiments}, we outline our experimental setup and present qualitative and quantitative results.

\section{Related Work}
\label{sec:related}
HTR can be tackled by considering different textual elements, \ie~ characters~\cite{jaderberg2015spatial,cilia2019ranking,clanuwat2019kuronet}, words~\cite{graves2009offline, bhunia2019handwriting, such2018fully,shi2016end}, lines~\cite{pham2014dropout, puigcerver2017multidimensional, voigtlaender2016handwriting, bluche2017gated, chowdhury2018efficient,moysset2017full, bluche2017scan}, paragraphs~\cite{bluche2017scan, bluche2016joint,wigington2018start}, or pages~\cite{clanuwat2019kuronet, moysset2017full}. Line-level HTR is among the most popular variants, which can be performed on pre-segmented text~\cite{pham2014dropout,voigtlaender2016handwriting,bluche2017gated,puigcerver2017multidimensional,chowdhury2018efficient}, or used in combination with layout analysis and line-level segmentation to obtain a paragraph-level or page-level HTR system~\cite{bluche2017scan,moysset2017full,yousef2020origaminet}. In this paper, we focus on pre-segmented line-level HTR.

Originally, HTR was tackled by applying Hidden Markov Models for image representation, and \emph{n}-gram based language models for textual output predictiont~\cite{toselli2004integrated,toselli2015handwritten}. The first deep learning-based solution to HTR was proposed in~\cite{graves2009offline}, where multi-dimensional Long Short-Term Memory networks (MDLSTM-RNNs) are used to build a 2D representation of the textual image, which is then collapsed in a sequence of vectors used for CTC decoding. This strategy was the standard one~\cite{bluche2016joint,moysset20192d,de2019no} until simpler alternatives to MDLSTM-RNNs were proposed~\cite{shi2016end,puigcerver2017multidimensional}. These consist of a ConvNet to extract a sequence of feature vectors from the text image and 1D-LSTMs to output character probabilities for the CTC decoding and became commonly used as a backbone for HTR systems~\cite{pham2014dropout, voigtlaender2016handwriting, bluche2017gated, chowdhury2018efficient,cojocaru2020watch} due to its performance and faster training compared to MDLSTMs.

A recent trend entails treating HTR as a sequence-to-sequence problem, from a sequence of text image slices encoded via convolutional and recurrent blocks to a sequence of transcribed text generated by a separate recurrent  block~\cite{sueiras2018offline, michael2019evaluating, zhang2019sequence, aberdam2021sequence}. Networks implementing this approach can be trained either by optimizing the cross-entropy loss alone or combined with the CTC loss. A variant of this approach entails using Transformers~\cite{vaswani2017attention} in place of RNNs~\cite{yousef2020origaminet,coquenet2020recurrence}, often requiring pre-training on either real or synthetic data ~\cite{kang2020pay,wick2021rescoring,li2021trocr} to obtain performance comparable to RNN-based solutions.
Finally, it is worth mentioning approaches that avoid using RNNs and are instead fully-convolutional~\cite{yousef2020origaminet,coquenet2020recurrence}. In these approaches, convolutional layers are combined with GateBlocks layers~\cite{yousef2018accurate} that operate a selection mechanism to model dependency similarly to LSTM cells. 

In the above-mentioned variants, recognition performance can be increased by integrating a language model, either a word-level or a character-level one. However, this strategy is effective only for those languages for which sufficient textual data are available. Thus, it is not always feasible for historical documents that are usually written in underrepresented languages. These can either be ancient versions of a modern language that has evolved over time or a language which is no longer spoken at all. In this work, we do not use any language model, both to assess the performance of the proposed HTR models in historical documents and to better enhance the benefits of deformable convolutions over standard convolutions.

Compared to OCR, HTR features the challenge related to the high variability of characters in shape and size. A common strategy is performing specific data augmentation and preprocessing~\cite{liu2005pseudo,poznanski2016cnn,voigtlaender2016handwriting,puigcerver2017multidimensional,wigington2017data,jayasundara2019textcaps, bera2019normalization, bouillon2019grayification}, while few works have faced this issue at the architectural design level. For example, in~\cite{zhong2016handwritten} a Spatial Transformer Network~\cite{jaderberg2015spatial} is employed for character-level HTR, while in~\cite{bhunia2019handwriting} an adversarial deformation module is used to warp intermediate convolutional features in a word-level HTR model.
In this paper, we propose to apply deformable convolutional kernels in the convolutional part of HTR models with the aim to tackle characters non-idealities without relying on data augmentation or specific preprocessing.


\section{Proposed Method}\label{sec:method}
In this section, we introduce our proposed approach. We first review deformable convolutions, and then introduce the convolutional-recurrent architectures we employ. 

\subsection{Deformable Convolutions}
ConvNets owe their success to their representational power and capacity to extract position-independent local features from images. Their key element is the convolutional operator  $*$, which performs a learned weighted sum of elements sampled over a regular grid $\mathcal{N}$. Formally, for a pixel $p$ of an input feature map $\bm{I}$, given a kernel $\bm{k}$ of learnable weights, the convolutional operator can be defined as follows:
\begin{equation}\label{eq:standard_conv}
    \left(\bm{I} * \bm{k}\right)(p) = \sum_{d \in \mathcal{N}} \bm{k}(d) \cdot \bm{I}(p + d),
\end{equation}
where $\cdot$ is the inner product between channel-wise feature vectors and $d$ is a displacement vector.
The size and structure of the sampling grid $\mathcal{N}$ depend on the kernel size and dilation. 

\begin{figure}[t]
\centering
\subfloat[]{
\includegraphics[width=0.3\columnwidth]{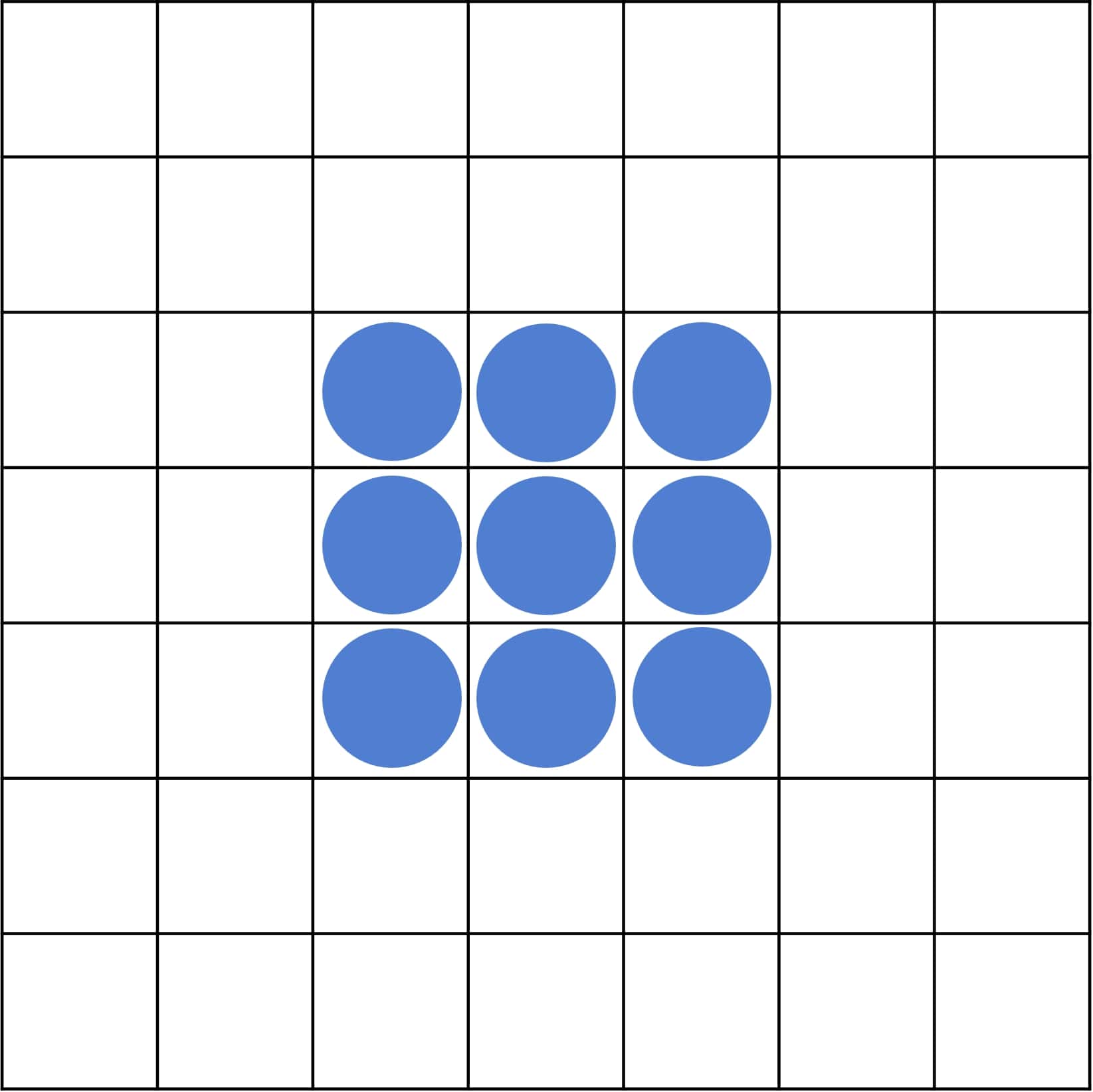}
\label{sfig:original}}
\subfloat[]{
\includegraphics[width=0.3\columnwidth]{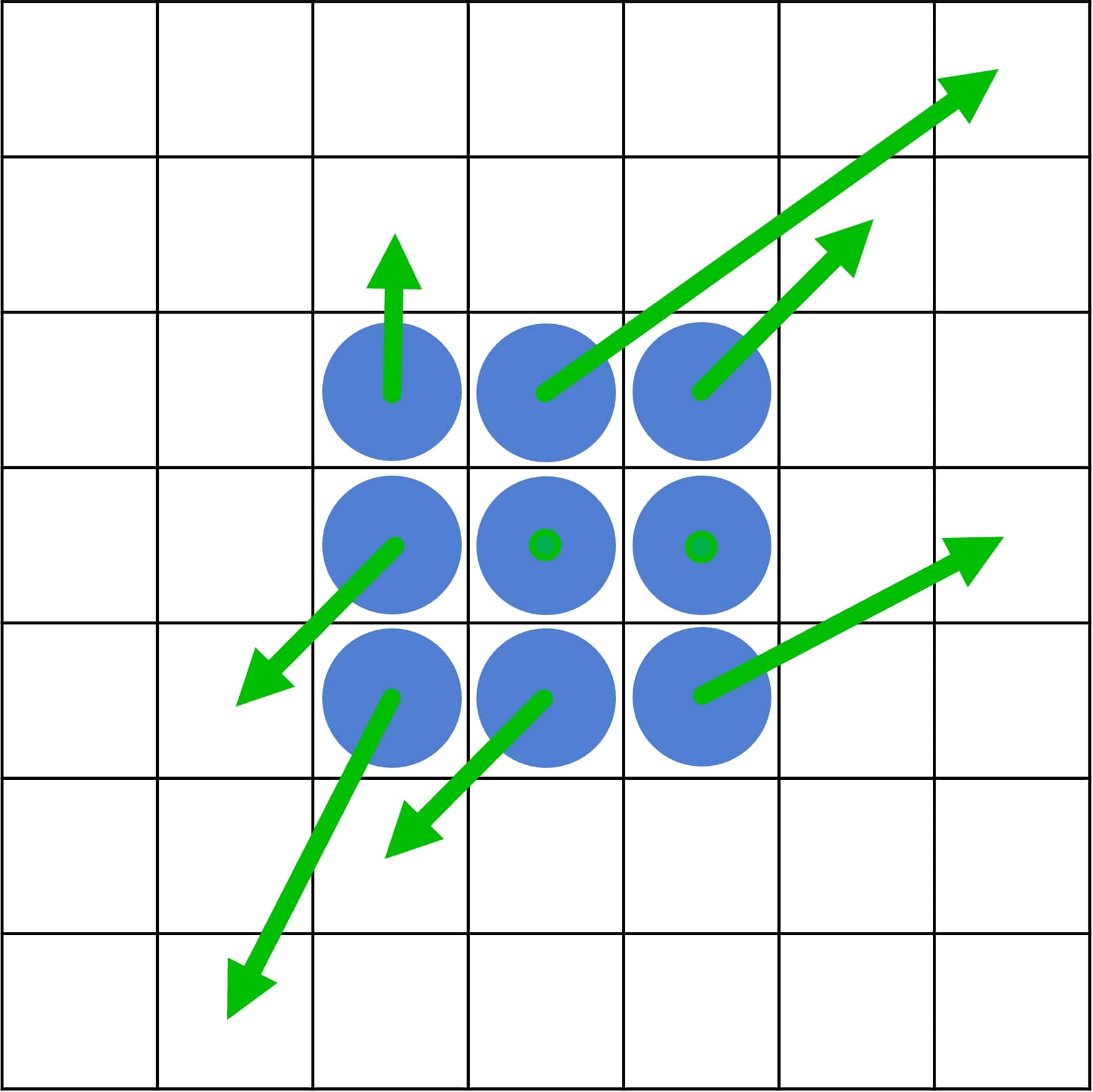}
\label{sfig:offsets}}
\subfloat[]{
\includegraphics[width=0.3\columnwidth]{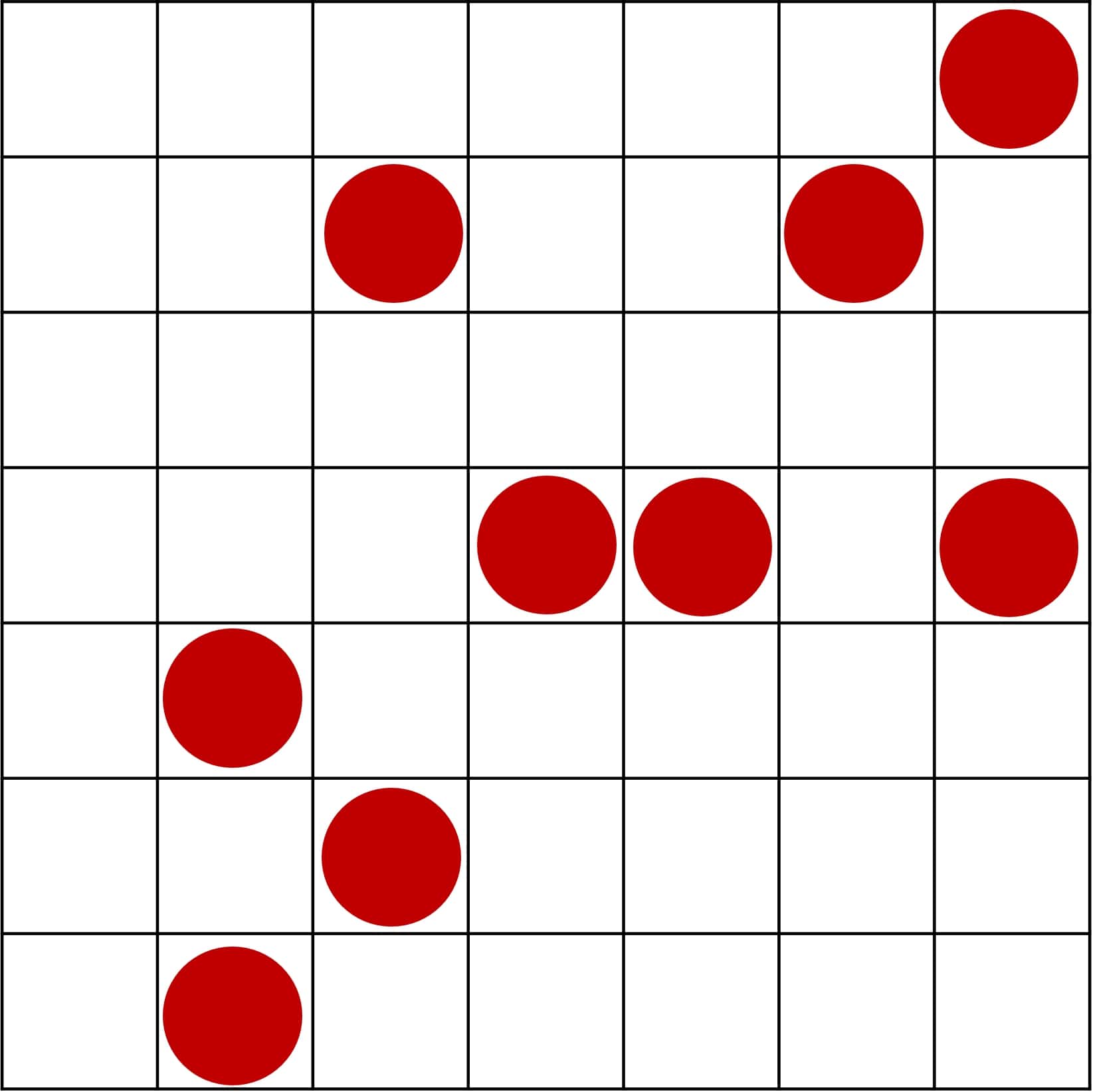}
\label{sfig:deformed}}
\caption{The regular sampling grid of a standard convolutional kernel (a) is deformed by applying a set of offsets (b), obtaining the deformed sampling grid of the deformable convolution (c).}
\label{fig:deformconv}
\end{figure}

Conversely, the deformable convolution operator $\circledast$~\cite{dai2017deformable} relies on an irregular sampling grid. The shape and geometry of the grid is learned as a function of the processed input context, which allows a content-dependent, non-regular feature extraction. To obtain the deformed grid, a regular sampling grid, such as that of a standard convolutional operator, is deformed by adding a learned 2D offset to each of its elements (as depicted in Figure~\ref{fig:deformconv}).

In practice, a deformable convolution layer is obtained by applying two standard convolutional layers: one is in charge of computing the offsets, the other of computing the kernel weights. Both are applied over the input feature map to ensure content-dependent deformation.
Formally, for deformable convolutions Eq.~\ref{eq:standard_conv} becomes 
\begin{equation}\label{eq:deformed_conv}
    \left(\bm{I} \circledast \bm{k}\right)(p) = \sum_{d \in \mathcal{N}} \bm{k}(d) \cdot \bm{I}(p + d + \delta(d)).
\end{equation}
Thus, the points included in the deformed kernel are those in the set $\{d + \delta(d)\}_{d \in \mathcal{N}}$.
Note that, in general, the computed offsets can be fractional numbers -- and simply quantizing them would harm the training phase. To overcome this limitation, Eq.~\ref{eq:deformed_conv} is implemented by including a 2D bilinear interpolation kernel $B(\cdot, \cdot)$, \ie
\begin{equation}\label{eq:deformed_conv_bilinear}
    \left( \bm{I} \circledast \bm{k} \right)(p) = \sum_{d \in \mathcal{N}} \bm{k}(d) \cdot \sum_{s \in \mathcal{S}} B(s, p + d + \delta(d)) \cdot \bm{I}(s),
\end{equation}
where $\mathcal{S}$ is the set of points in $\bm{I}$ that are in the neighborhood of the sampling locations $\{p + d+ \delta(d)\}_{d \in \mathcal{N}}$.

\begin{figure}[t]
\centering
\includegraphics[width=\columnwidth]{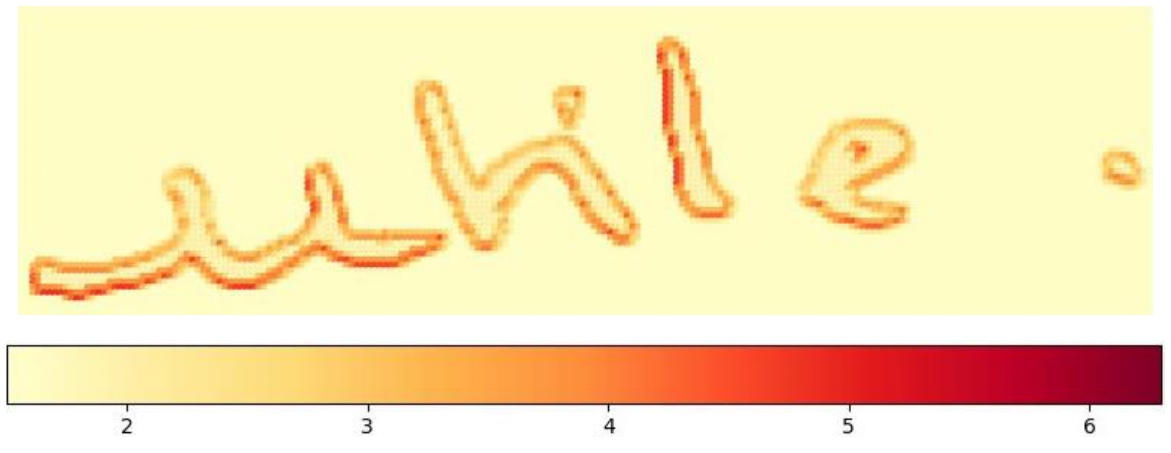}
\caption{Cumulative magnitude of the offsets applied to a $3\times3$ kernel grid on points of an image of a word (first convolutional layer of CRNN). Grids sampling in uniform regions are less deformed than those sampling on edges.}
\label{fig:offset_mag}
\end{figure}

To appreciate the effects of the deformable convolution on handwriting images, in Figure~\ref{fig:overview} we represent some deformed sampling grids obtained from a $3\times3$ deformable kernel applied to the image of a character, in comparison with those of a standard covolutional kernel. It can be noticed that when the kernel is applied to stroke edges it is considerably deformed to adapt to its input. On the other hand, on uniform regions such as background and solid ink, the deformation is less evident. The same behavior can be observed from Figure~\ref{fig:offset_mag}, which represents the cumulative magnitudes of the offsets of a $3\times3$ deformable kernel applied to all the pixels of a short text line. When processing pixels from uniform areas, the learned offsets are small, thus the kernel does not deform much, while are larger when processing pixels on the edges.

Given a deformable convolutional layer with square kernels, kernel size $k$, $c_{in}$ input channels, and $c_{out}$ output channels, the number of weights needed for computing the deformable convolution is $k^2 \cdot c_{in} \cdot c_{out}$\footnote{For the sake of simplicity, we do not include biases in this analysis.}, which is comparable to those needed by a standard convolutional layer. If the convolutional layer computing the offsets is also created with the same kernel size, it will add $k^2 \cdot c_{in} \cdot (2 k^2)$ parameters, for a total of $k^2 \cdot c_{in} \cdot (2 k^2 + c_{out})$ weights. From the point of view of the memory footprint, therefore, replacing a standard convolutional layer with a deformable one implies the same cost of adding $2 k^2$ output feature maps.

\begin{figure*}[t]
\centering
\includegraphics[width=\linewidth]{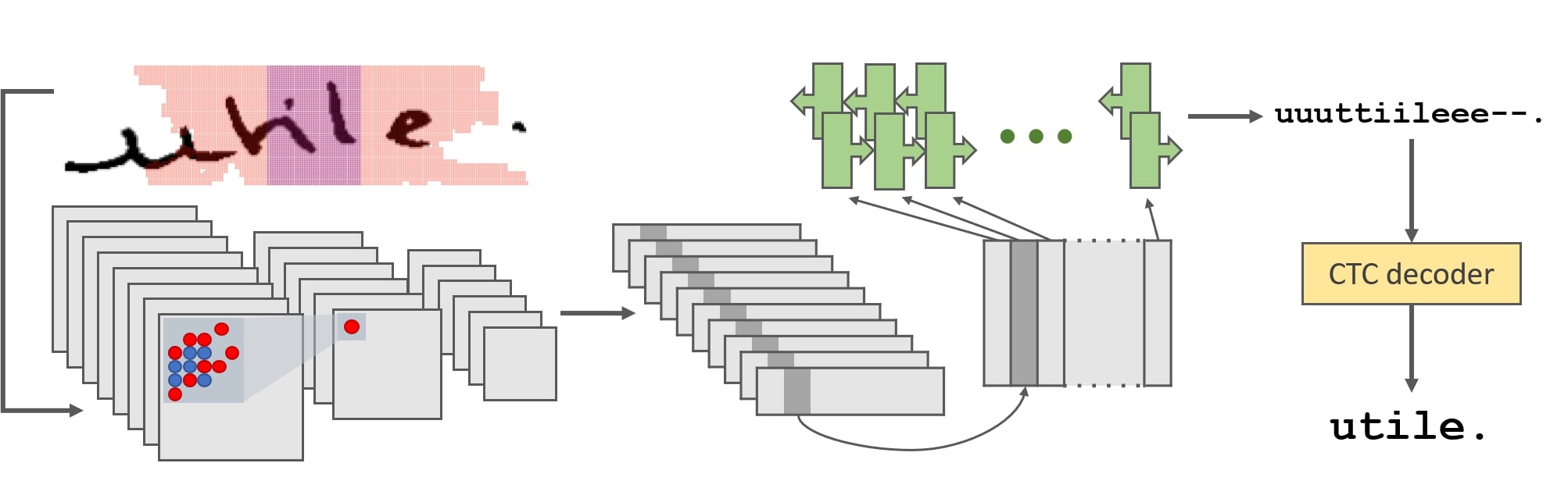}
\caption{General scheme of a convolutional-recurrent HTR system featuring deformable convolutions.}
\label{fig:general_system}
\end{figure*}

\subsection{Deformable Convolutional-Recurrent Network}
Given the capability of deformable convolutions to adapt to geometric transformations, we propose to employ them instead of regular convolutions in HTR architectures. To this end, we build upon two commonly used backbone models for HTR: the sequence recognition network proposed in~\cite{shi2016end} and the model presented in~\cite{puigcerver2017multidimensional}. In the following, we refer to these models as CRNN and 1D-LSTM, respectively.

Both models take as input an image representing a row of handwritten text and consist of three main components: a ConvNet to extract visual features from the input image, an RNN which treats the visual feature map as a sequence and outputs character probabilities, and a decoding block to output the final transcription. 
For training the models, we maximize the CTC probability of the output sequence. Thus, additional to textual characters, the RNN scores a special \emph{blank} character that means ``no other character''.
An overview of the main components of these HTR systems is shown in Figure~\ref{fig:general_system}.

The convolutional part of CRNN has the same architecture as of VGG-11 \cite{simonyan2014very} up to the sixth convolutional block, to which we add another convolutional block with a $2\times2$ kernel. Additionally, the receptive field of the 3\textsuperscript{rd} and 4\textsuperscript{th} max-pooling layers are changed from squared $2\times2$ into rectangular $2\times1$ in order to obtain wider feature maps reflecting the common height-width ratio of images of text lines. Note that in this architecture, all the convolutional layers are deformable.

For the convolutional part of 1D-LSTM, we stack five blocks containing a deformable convolution layer with $3 \times 3$ kernels, a Batch Normalization layer, and a LeakyReLU activation function. The deformable convolution layer of the first block has 16 filters, and for the others, we increase the number of filters by 16 at each block. We apply $2 \times 2$ max-pooling to the output of the first three blocks and leave the output of the last two blocks as it is. 

\begin{figure*}[t]
\centering
\footnotesize
\setlength{\tabcolsep}{.5em}
\resizebox{\linewidth}{!}{
\begin{tabular}{ccccc}
\multicolumn{2}{c}{\textbf{IAM dataset}} & & \multicolumn{2}{c}{\textbf{RIMES dataset}} \\
\addlinespace[0.12cm]
\multicolumn{2}{c}{\includegraphics[width=0.5\linewidth]{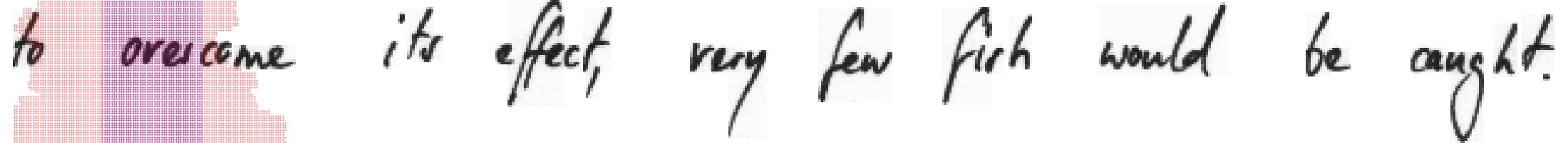}} & & \multicolumn{2}{c}{\includegraphics[width=0.5\linewidth]{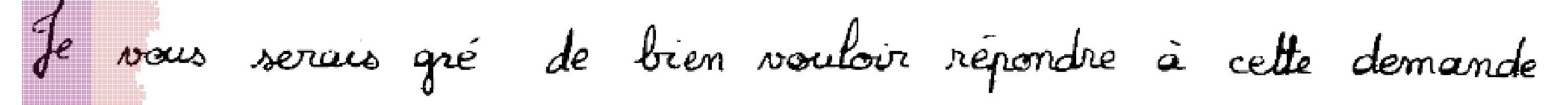}} \\
\multicolumn{2}{c}{\includegraphics[width=0.5\linewidth]{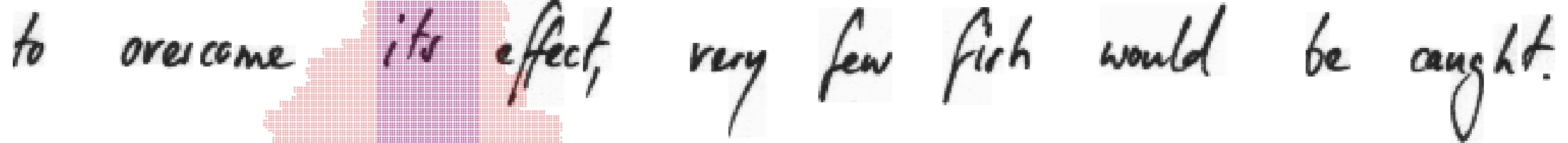}} & & \multicolumn{2}{c}{\includegraphics[width=0.5\linewidth]{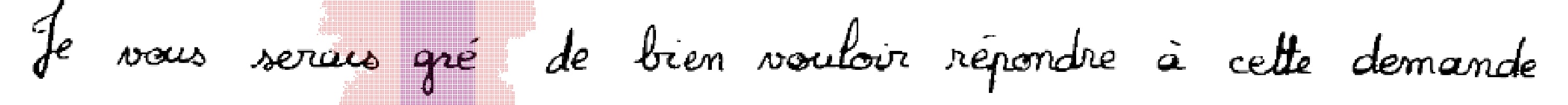}} \\
\multicolumn{2}{c}{\includegraphics[width=0.5\linewidth]{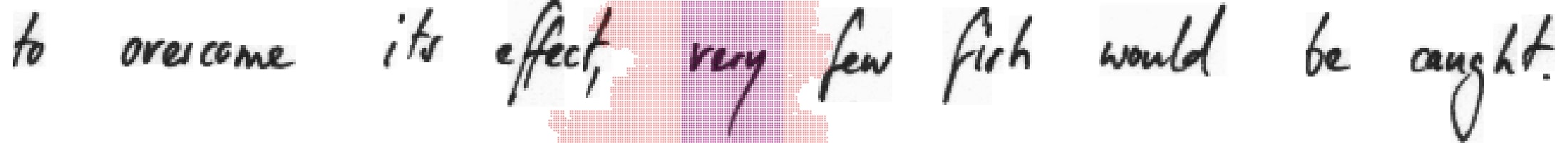}} & & \multicolumn{2}{c}{\includegraphics[width=0.5\linewidth]{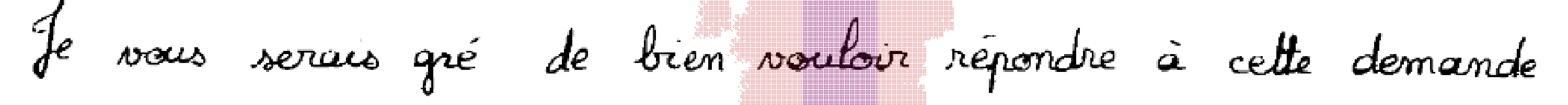}} \\
\multicolumn{2}{c}{\includegraphics[width=0.5\linewidth]{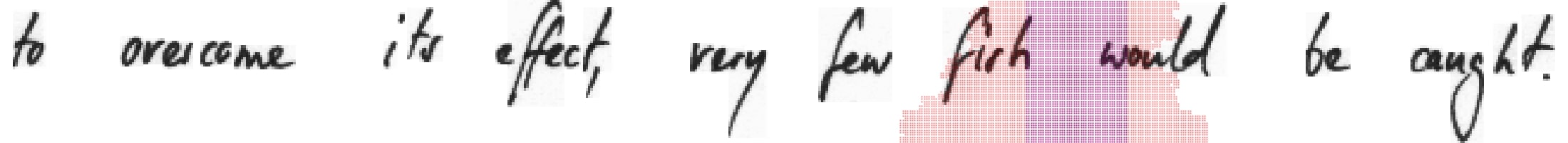}} & & \multicolumn{2}{c}{\includegraphics[width=0.5\linewidth]{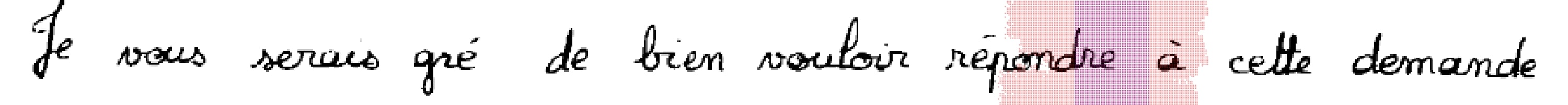}} \\
\multicolumn{2}{c}{\includegraphics[width=0.5\linewidth]{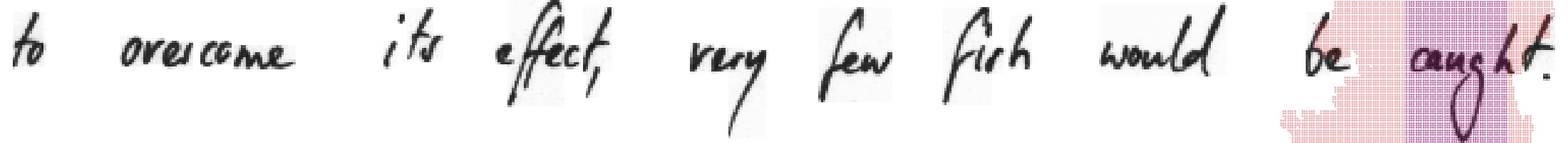}} & & \multicolumn{2}{c}{\includegraphics[width=0.5\linewidth]{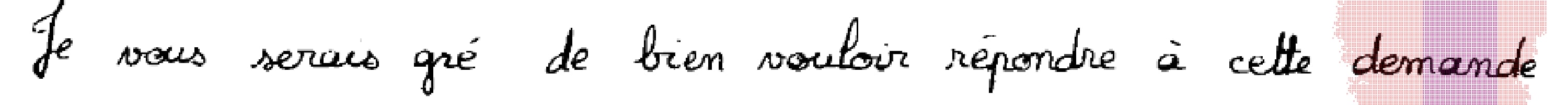}} \\
\addlinespace[0.12cm]
\addlinespace[0.12cm]
\multicolumn{2}{c}{\textbf{ICFHR14 dataset}} & & \multicolumn{2}{c}{\textbf{Leopardi dataset}} \\
\addlinespace[0.12cm]
\multicolumn{2}{c}{\includegraphics[width=0.5\linewidth]{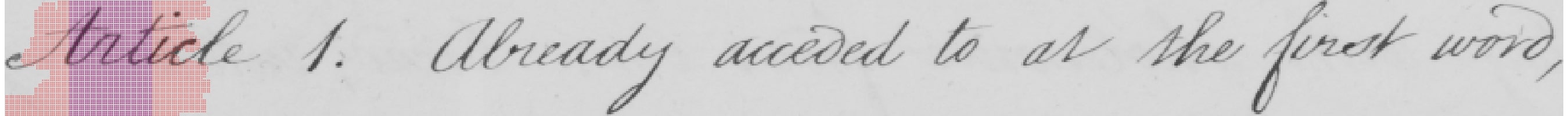}} & & \multicolumn{2}{c}{\includegraphics[width=0.5\linewidth]{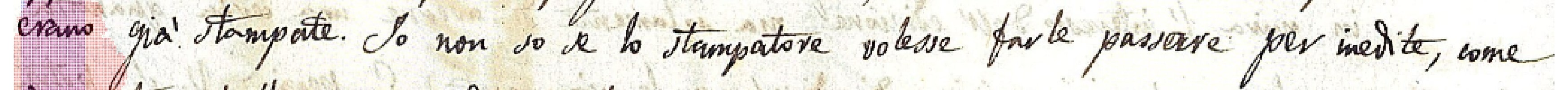}} \\
\multicolumn{2}{c}{\includegraphics[width=0.5\linewidth]{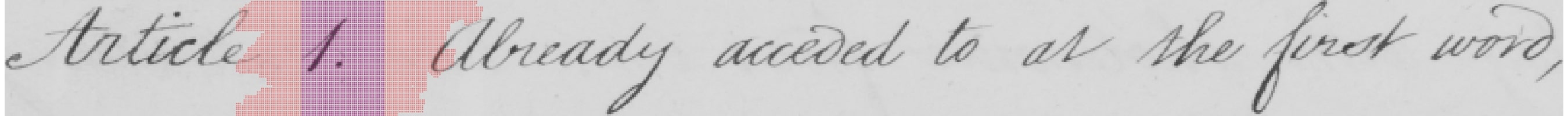}} & & \multicolumn{2}{c}{\includegraphics[width=0.5\linewidth]{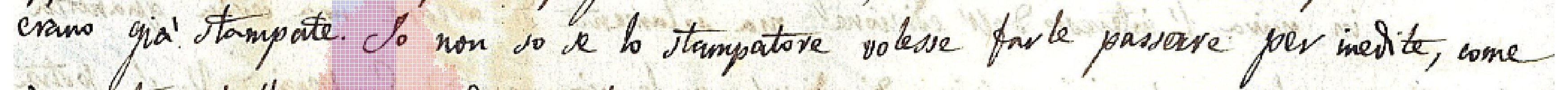}} \\
\multicolumn{2}{c}{\includegraphics[width=0.5\linewidth]{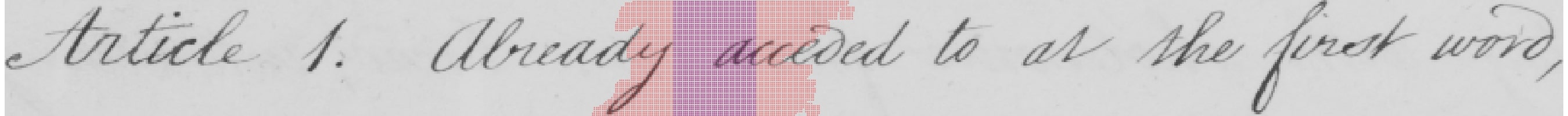}} & & \multicolumn{2}{c}{\includegraphics[width=0.5\linewidth]{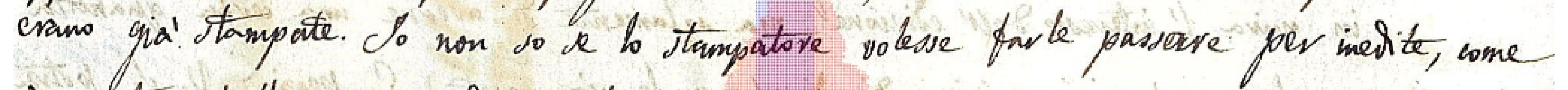}} \\
\multicolumn{2}{c}{\includegraphics[width=0.5\linewidth]{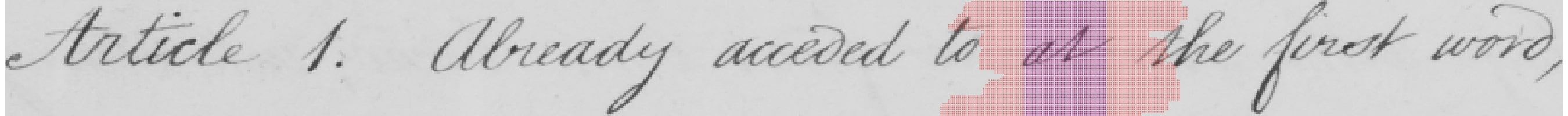}} & & \multicolumn{2}{c}{\includegraphics[width=0.5\linewidth]{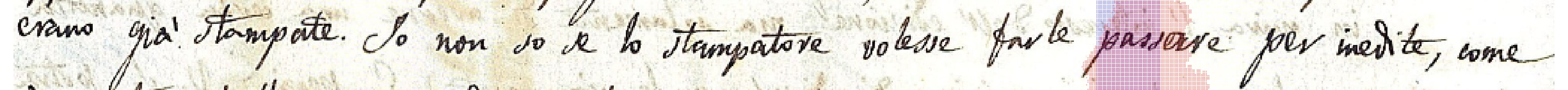}} \\
\multicolumn{2}{c}{\includegraphics[width=0.5\linewidth]{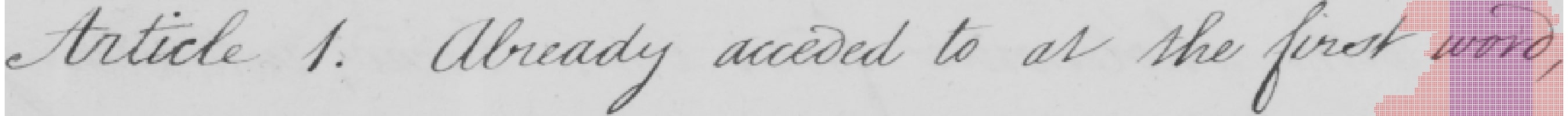}} & & \multicolumn{2}{c}{\includegraphics[width=0.5\linewidth]{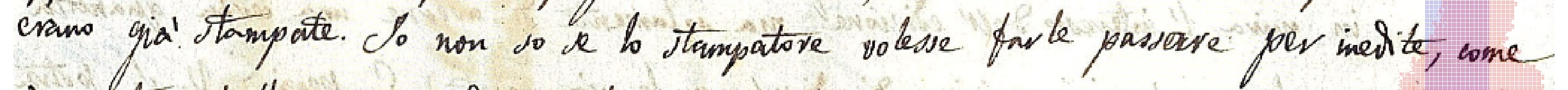}} \\
\end{tabular}
}
\caption{Some receptive fields of the HTR network using standard convolutions (in transparent blue) and deformable convolutions (in transparent red) on modern (top) and historical (bottom) text line images. Deformable convolutions lead to areas of irregular shape that better adapt to handwritten strokes and cover a wider portion of the image thanks to the limited amount of additional offsets parameters (best seen in color).}
\label{fig:receptive_fields}
\end{figure*}

In both the proposed variants, the $H\times W\times C$ feature map of the last convolutional layer is used to obtain a sequence of $W$ $(H \cdot C)$-elements feature vectors that serve as input for the recurrent part. These feature vectors are obtained by concatenating the $C$-dimensional vectors on the $H$ rows of the map and represent regions of the original image, \ie~the receptive field. Due to the deformable kernels, such receptive fields have irregular shape and can possibly be non-connected, which allows them to cover a wider portion of the input and better adapt to its patterns (see some examples reported in Figure~\ref{fig:receptive_fields}).

For the recurrent part of both the CRNN and 1D-LSTM variants we use Bidirectional LSTMs (BLSTMs). We stack two BLSTMs layers in the CRNN variant and five in the 1D-LSTM. At each timestep, the recurrent part takes as input a feature vector in the sequence obtained from the last convolutional feature map, from left to right, and outputs the probabilities of each character to be in the corresponding image region.

Finally, in the decoding block, the transcription is obtained via greedy decoding, \ie~by concatenating the labels with highest probability at each timestep. Then, all the duplicate labels that are not separated by the \emph{blank} token are collapsed in a single character and all the \emph{blank} tokens are removed.
The detailed architectures of the presented CRNN and 1D-LSTM models are reported in the Appendix.


\section{Experimental Evaluation}\label{sec:experiments}
In this section, we evaluate the suitability of the proposed method for the HTR task based on deformable convolutions (DefConvs) when compared to baselines that feature standard convolutions (StdConvs). For clarity, in this section, we refer to our proposed models as DefConv CRNN and DefConv 1D-LSTM, and we refer to their counterparts containing only standard convolutions as StdConv CRNN and StdConv 1D-LSTM, respectively. 

\subsection{Datasets}
In our experiments, we first consider the IAM~\cite{marti2002iam} and the RIMES~\cite{augustin2006rimes} datasets, which are commonly used as benchmark for line-level HTR. Both datasets feature different handwriting styles due to the presence of multiple writers and contain samples with curved text lines. Moreover, to validate the ability to deal with uneven background and other nuisances typical of aged manuscripts, we also use two benchmark line-level historical datasets, namely the ICFHR14~\cite{sanchez2014icfhr2014} and ICFHR16~\cite{sanchez2016icfhr2016} datasets. Finally, to test the benefits of DefConvs in a pre-training plus fine-tuning setting for HTR on small historical manuscripts~\cite{aradillas2020boosting}, we consider the recently proposed Leopardi dataset~\cite{cascianelli2021learning}. 

\tit{IAM} The IAM Handwriting dataset contains free-layout modern English text lines from the Lancaster-Oslo/Bergen (LOB) corpus~\cite{johansson1978manual}, handwritten by multiple users. Following the commonly adopted Aachen University splitting\footnote{\url{www.tbluche.com/resources.html}}, the dataset consists of $6\,482$ training lines, $976$ validation lines, and $2\,915$ test lines. Non-\emph{blank} characters are $95$, and the line images are $1698\pm292$ pixels wide and $124\pm34$ pixels high.

\tit{RIMES} The RIMES dataset contains unconstrained modern French letters handwritten by multiple users. In our experiments, we consider the official train/test splitting consisting of $11\,333$ training lines and $778$ test lines and obtain the validation set by retaining the lines contained in the $10\%$ of the training documents. The total number of non-\emph{blank} characters in this dataset is $79$, and the line images width and height are $1637\pm555$ pixels and $130\pm36$ pixels, respectively. 

\tit{ICFHR14} The ICFHR14 dataset features legal forms and drafts from the Bentham Papers collection~\cite{causer2012building}, handwritten by the English philosopher and renovator Jeremy Bentham and his collaborators from mid-18\textsuperscript{th} century to mid-19\textsuperscript{th} century. The dataset was used in a competition for the ICFHR conference in 2014, from which we use the indicated splitting\footnote{\url{http://doi.org/10.5281/zenodo.44519}} consisting of $9\,198$ lines for training, $1\,415$ for validation, and $860$ for test. The total number of non-\emph{blank} characters in this dataset is $88$, and the line images width and height are $1401\pm573$ pixels and $121\pm38$ pixels, respectively. 

\tit{ICFHR16} The ICFHR16 dataset features minutes of council meetings from the Ratsprotokolle collection, handwritten by multiple writers from 1470 to 1805 in old German~\cite{sanchez2016icfhr2016}. The dataset was used in a competition for the ICFHR conference in 2016\footnote{\url{http://doi.org/10.5281/zenodo.1164045}}. In this dataset, there are $8\,367$ lines for training, $1\,043$ for validation, and $1\,140$ for test. Non-\emph{blank} characters in this dataset are $93$, and the images contained are $963\pm318$ pixels wide, and $123\pm39$ pixels high.

\tit{Leopardi} The Leopardi dataset consists of a small collection of early 19\textsuperscript{th} Century letters written in Italian by Giacomo Leopardi~\cite{cascianelli2021learning}. It contains $1\,303$ training lines, $587$ validation lines, and $569$ test lines. The total number of non-\emph{blank} characters in this dataset is $77$, and the images contained are $1597\pm593$ pixels wide, and $124\pm30$ pixels high. This dataset comes with synthetic data that can be used for pre-training plus fine-tuning. The synthetic data are divided in two sets: in one the handwriting resembles that of the original author on historical manuscripts, in the other, the handwriting is modern. The text is obtained from some Leopardi's proses, so that the language is contemporary to that used in the letters contained in the original Leopardi dataset. Both synthetic sets consists of $89\,068$ training lines and $22\,397$ validation lines, and non-\emph{blank} characters are $114$.

\subsection{Implementation Details}
As the sole pre-processing steps, we normalize the text line images between $-1$ and $1$ and rescale them in height, keeping the original aspect ratio. In particular, we rescale them to become $60$ pixels high for the CRNN model, $128$ for the 1D-LSTM model.
The output of the convolutional component is the feature map of its last layer, which is a $2\times W \times 512$ tensor in the CRNN model and a $16\times W \times 80$ tensor in the 1D-LSTM model. These are collapsed in a sequence of $W$ vectors of $1024$ and $1280$ elements, respectively.
The BLSTMs that constitute the recurrent part of the models have $512$ hidden units each in the CRNN variant and $256$ in the 1D-LSTM variant. In both cases, the recurrent layers are separated by a dropout layer with dropout probability equal to $0.5$. 
The proposed models have been trained with batch size equal to $8$ for the CRNN variant and $2$ for the 1D-LSTM variant using Adam as optimizer with $\beta_1 = 0.9$ and $\beta_2 = 0.999$, and learning rate equal to $0.0001$ for the CRNN variant and to $0.003$ for the 1D-LSTM variant. 
We train the models until the Character Error Rate (CER) on the validation set stops improving for $20$ epochs. Further details on the models architecture can be found in the Appendix.

For the pretraining plus fine-tuning experiment on the Leopardi dataset, we use the two available synthetic sets separately for pre-training the StdConv and DefConv-based variants. In this phase, we apply random distortions to alter the lines appearance and shape, as done in~\cite{cascianelli2021learning}. The batch size is set to $16$ and the learning rate to $0.0001$. Also in this case, we stop the training when the validation CER does not improve for $20$ epochs. After pre-training, we fine-tune on subsets of decreasing number of training lines from the original Leopardi dataset. 

\begin{table}[t]
\centering
\small
\setlength{\tabcolsep}{.55em}
\caption{Results on the IAM dataset and the RIMES dataset. Note that $^\dagger$ and $^\ddagger$ indicate results of re-implemented method as from~\cite{moysset20192d} and~\cite{puigcerver2017multidimensional}, respectively, without language model.}
\label{tab:mod_results}
\resizebox{\columnwidth}{!}{%
\begin{tabu}{lcccccc}
\toprule
      & & \multicolumn{2}{c}{\textbf{IAM}} & & \multicolumn{2}{c}{\textbf{RIMES}} \\
\cmidrule{3-4}
\cmidrule{6-7}
\textbf{Method}         & & \textbf{CER} & \textbf{WER} & & \textbf{CER} & \textbf{WER}        \\
\midrule
\textbf{de Buy Wenniger~\etal~\cite{de2019no}}              & &  12.9 & 40.8 & & -   & -    \\ 
\textbf{Chen~\etal~\cite{chen2017simultaneous}}                                  & &  11.2 & 34.6 & & 8.3 & 30.5 \\ 
\textbf{Pham~\etal~\cite{pham2014dropout}}                                       & &  10.8 & 35.1 & & 6.8 & 28.5 \\ 
\textbf{Bluche and Messina~\cite{bluche2017gated}$^\dagger$} & &  10.2 & 32.9 & & 5.8 & 19.7\\
\textbf{Moysset and Messina~\cite{moysset20192d}}           & &   8.9 & 29.3 & & 4.8 & 16.4 \\ 
\textbf{Zhang~\etal~\cite{zhang2019sequence}}               & &   8.5 & 22.2 & & -   & -    \\ 
\textbf{Voigtlaender~\etal~\cite{voigtlaender2016handwriting}$^\ddagger$}                  & &   8.3 & 27.5 & & 4.0 & 17.7 \\ 
\textbf{Chowdhury and Vig~\cite{chowdhury2018efficient}}    & &   8.1 & 16.7 & & 3.5 & 9.6  \\ 
\textbf{Coquenet~\etal~\cite{coquenet2020recurrence}}       & &   8.0 & 28.6 & & 4.4 & 18.0 \\ 
\textbf{Bluche~\cite{bluche2016joint}}                      & &   7.9 & 24.6 & & 2.9 & 12.6 \\ 
\textbf{Markou~\etal~\cite{markou2021convolutional}}        & &   7.9 & 23.9 & & 3.9 & 13.4 \\ 
\textbf{Kang~\etal~\cite{kang2020pay}}                      & &   7.6 & 24.5 & & -   & -    \\ 
\textbf{Ly~\etal~\cite{ly2021self}}                         & &   7.2 & 22.9 & & -   & -    \\ 
\midrule
\textbf{StdConv 1D-LSTM}                                                         & &  7.7         & \textbf{26.3} & & 5.8 & 25.5 \\
\textbf{DefConv 1D-LSTM}                                                         & & \textbf{7.5} & 26.9          & & \textbf{5.2} & \textbf{23.7} \\
\midrule
\textbf{StdConv CRNN}                                                            & &  7.8         & 27.8          & & 4.4 & 16.0 \\
\textbf{DefConv CRNN}                                                            & & \textbf{6.8} & \textbf{24.7} & & \textbf{4.0} & \textbf{13.7} \\
\bottomrule
\end{tabu}
}
\end{table}

\begin{table}[t]
\centering
\small
\setlength{\tabcolsep}{.25em}
\caption{Results on the ICFHR14, ICFHR16, and Leopardi datasets.}
\label{tab:hist_results}
\resizebox{\columnwidth}{!}{%
\begin{tabular}{lccccccccc}
\toprule & & \multicolumn{2}{c}{\textbf{ICFHR14}} & & \multicolumn{2}{c}{\textbf{ICFHR16}} & & \multicolumn{2}{c}{\textbf{Leopardi}} \\
\cmidrule{3-4}
\cmidrule{6-7}
\cmidrule{9-10}
\textbf{Method}             & & \textbf{CER} & \textbf{WER} &      & \textbf{CER} & \textbf{WER} &     & \textbf{CER} & \textbf{WER}\\
\midrule
\textbf{StdConv 1D-LSTM}    & & 4.8          & 15.3          &     & 5.8           & 25.5            &    & 3.8            & 13.8          \\
\textbf{DefConv 1D-LSTM}    & & \textbf{3.6} & \textbf{14.3} &     & \textbf{5.2}  & \textbf{23.7}   &    & \textbf{3.4}   & \textbf{12.6} \\
\midrule
\textbf{StdConv CRNN}       & & 3.9          & 15.3          &     & 5.5           & 25.9            &    & 3.4            & 13.4          \\
\textbf{DefConv CRNN}       & & \textbf{3.6} & \textbf{13.9} &     & \textbf{4.5}  & \textbf{21.7}   &    & \textbf{2.8}   & \textbf{10.8} \\
\bottomrule
\end{tabular}
}
\end{table}

\begin{figure*}[t]
\centering
\footnotesize
\setlength{\tabcolsep}{.7em}
\resizebox{\linewidth}{!}{
\begin{tabular}{llcll}
\multicolumn{5}{c}{\textbf{IAM dataset}} \\
\multicolumn{2}{c}{\includegraphics[width=0.45\linewidth]{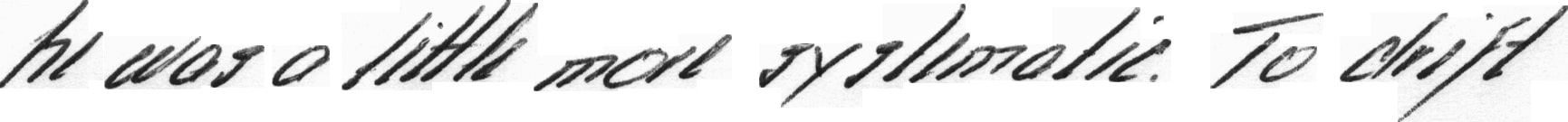}} & & \multicolumn{2}{c}{\includegraphics[width=0.45\linewidth]{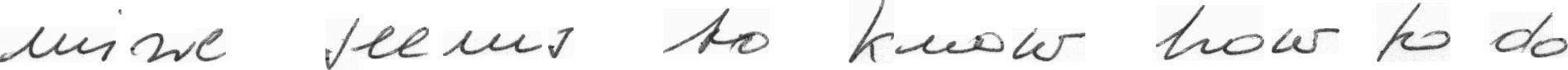}} \\
\textbf{Ground Truth}         & he was a little more systematic. To drift & & 
\textbf{Ground Truth}         & mine seems to know how to do \\
\cmidrule{1-2} \cmidrule{4-5}
\textbf{StdConv 1D-LSTM} & he was o litlle more syshmolica. To chif & & 
\textbf{StdConv 1D-LSTM} &  iave teems to krmour hour to do \\
\textbf{DefConv 1D-LSTM}      & he was a little more syslumalics. To diff & & 
\textbf{DefConv 1D-LSTM}      & uiove seems to kmow how to do \\
\textbf{StdConv CRNN}    & hI wos o titll morl syslmalics. To chift & & 
\textbf{StdConv CRNN}    &  miare Jeemns tro kwow howr to do \\
\textbf{DefConv CRNN}         & he war o little more systmatie. To drift & & 
\textbf{DefConv CRNN}         &  miare seemns tro knwow how to do\\
\addlinespace[0.24cm]
\multicolumn{5}{c}{\textbf{RIMES dataset}} \\
\multicolumn{2}{c}{\includegraphics[width=0.45\linewidth]{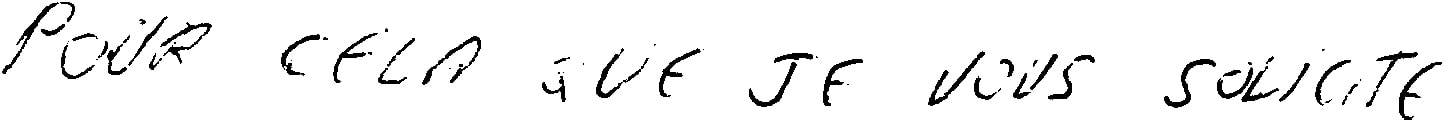}} & & \multicolumn{2}{c}{\includegraphics[width=0.45\linewidth]{images/qualitatives/eval2011-39-3.jpg}} \\
\textbf{Ground Truth}         & POUR CELA QUE JE VOUS SOLICITE & & 
\textbf{Ground Truth}         & au nom de ORY Pascale (Référence client : EXOXU51). \\
\cmidrule{1-2} \cmidrule{4-5}
\textbf{StdConv 1D-LSTM} & Pair CLA IUE JE VUS SOUETE & & 
\textbf{StdConv 1D-LSTM} & au nom de OPy Pascale (référence clientiGrOnSN). \\
\textbf{DefConv 1D-LSTM}      & POULR CELD SUE JE VOUS SOLIETE  & & 
\textbf{DefConv 1D-LSTM}      & au nom de ORy Pascale (référence clientEXONUS1). \\
\textbf{StdConv CRNN}    & PoIiR CELA IVE JE VOUS SOUICITE & & 
\textbf{StdConv CRNN}    & au nom de ORy Poscale (référence clientiFXoXUS1). \\
\textbf{DefConv CRNN}         & POUR CFLA IUE JE VOUS SOLITE & & 
\textbf{DefConv CRNN}         & au nom de Ory Pascale (référence client : EXOXU51). \\
\end{tabular}
}
\caption{Qualitative results on the benchmark modern datasets considered.}
\label{fig:modern_qualitatives}
\end{figure*}

\subsection{Compared Approaches}
We perform a direct comparison of our models and the corresponding baselines not featuring DefConvs on all the considered datasets. For the baselines, we apply the same design choices and training strategies as for our proposed DefConv-based models. Moreover, for the IAM and RIMES datasets, we report the results of other approaches in literature that exploit standard convolution and not perform any data augmentation or language model correction. This way, we can analyse more clearly the effects of the deformable convolution with respect to the standard one. In particular, we consider the approaches proposed by Pham \etal~\cite{pham2014dropout}, Voigtlaender \etal~\cite{voigtlaender2016handwriting}, Chen \etal~\cite{chen2017simultaneous}, de Buy Wenniger \etal~\cite{de2019no}, Moysset and Messina~\cite{moysset20192d}, and Bluche~\cite{bluche2016joint},
which are based on the MDLSTM-RNN network proposed in~\cite{graves2009offline}. Moreover, we include in the analysis approaches employing 1D-LSTMs, such as those presented by Ly \etal~\cite{ly2021self}, which also exploits self-attention in the convolutional block, and Markou \etal~\cite{markou2021convolutional}, which features a fully-connected layer after the recurrent block. We also consider the fully-convolutional architectured by Bluche and Messina~\cite{bluche2017gated}, and Coquenet \etal~\cite{coquenet2020recurrence}. Finally, we compare against approaches following the sequence-to-sequence paradigm, such as those proposed by Zhang \etal~\cite{zhang2019sequence}, Chowdhury and Vig~\cite{chowdhury2018efficient}, and Kang \etal~\cite{kang2020pay}, this latter based on Transformers.

\subsection{Results}

The obtained results are summarized in Table~\ref{tab:mod_results} for the modern datasets and Table~\ref{tab:hist_results} for the historical datasets. The performance is expressed in terms of the commonly used metrics, CER and the Word Error Rate (WER).

\begin{table*}[t]
\centering
\small
\setlength{\tabcolsep}{.35em}
\caption{Fine-tuning results at varying number of fine-tuning lines.}
\label{tab:fine-tuning}
\resizebox{\textwidth}{!}{%
\begin{tabular}{lccc ccccccccccccccccc}
\toprule 
& & & & \multicolumn{2}{c}{\textbf{100\%}} & & \multicolumn{2}{c}{\textbf{50\%}} & & \multicolumn{2}{c}{\textbf{25\%}}  & & \multicolumn{2}{c}{\textbf{10\%}} & & \multicolumn{2}{c}{\textbf{5\%}} & & \multicolumn{2}{c}{\textbf{2,5\%}} \\
\cmidrule{5-6} \cmidrule{8-9} \cmidrule{11-12} \cmidrule{14-15} \cmidrule{17-18} \cmidrule{20-21}
\textbf{Method} & & \textbf{Pre-training} & & \textbf{CER} & \textbf{WER} & & \textbf{CER} & \textbf{WER} & & \textbf{CER} & \textbf{WER} & & \textbf{CER} & \textbf{WER} & & \textbf{CER} & \textbf{WER} & & \textbf{CER} & \textbf{WER}\\
\midrule
\textbf{StdConv 1D-LSTM} & & modern & & 2.9 & 10.9 & & 4.4 & 16.1 & & 7.6 & 25.5 & & 13.6 & 40.0 & & 18.1 & 49.6 & & 90.8 & 99.4 \\
\textbf{DefConv 1D-LSTM} & & modern & & \textbf{2.5} &  \textbf{9.3} & & \textbf{4.2} & \textbf{15.5} & & \textbf{5.7} & \textbf{20.6} & & \textbf{13.0} & \textbf{38.0} & & \textbf{17.1} & \textbf{48.5} & & \textbf{26.6} & \textbf{65.5} \\
\midrule
\textbf{StdConv CRNN}    & & modern & & 3.1 & 12.3 & & 4.9 & 19.0 & & 7.6 & 27.4 & & 11.8 &   38.0  & & 14.8 & 45.1 & & 20.0 & 56.4 \\
\textbf{DefConv CRNN}    & & modern & & \textbf{2.6} & \textbf{10.0} & & \textbf{3.5} & \textbf{13.7} & & \textbf{5.9} & \textbf{21.7} & &  \textbf{9.8} & \textbf{32.6} & & \textbf{12.8} & \textbf{41.0} & & \textbf{17.7} & \textbf{52.0} \\
\midrule
\textbf{StdConv 1D-LSTM} & & historical & & 2.5 & 9.2 & & 4.3 & 16.1 & & 6.4 & 22.4 & & 9.9 & 30.7 & & 12.1 & 37.5 & & 16.8 & 47.4 \\
\textbf{DefConv 1D-LSTM} & & historical & & \textbf{2.3} & \textbf{8.7} & & \textbf{3.5} & \textbf{13.4} & & \textbf{5.0} & \textbf{18.5} & & \textbf{8.7} & \textbf{27.6} & &  \textbf{9.6} & \textbf{31.4} & & \textbf{15.0} & \textbf{44.9} \\
\midrule
\textbf{StdConv CRNN}    & & historical & & 2.7 & 10.8 & & 4.2 & 16.2 & & 6.2 & 22.6 & & 9.3 & 31.8 & & 11.7 & 38.0 & & 14.4 & 44.9 \\
\textbf{DefConv CRNN}    & & historical & & \textbf{2.3} &  \textbf{8.9} & & \textbf{3.5} & \textbf{13.8} & & \textbf{5.1} & \textbf{19.2} & & \textbf{7.8} & \textbf{27.7} & &  \textbf{9.4} & \textbf{32.4} & & \textbf{13.5} & \textbf{44.0} \\
\bottomrule
\end{tabular}
}
\end{table*}

\begin{figure*}[t]
\centering
\footnotesize
\setlength{\tabcolsep}{.7em}
\resizebox{\linewidth}{!}{
\begin{tabular}{llcll}
\multicolumn{5}{c}{\textbf{ICFHR14 dataset}}\\
\multicolumn{2}{c}{\includegraphics[width=0.45\linewidth]{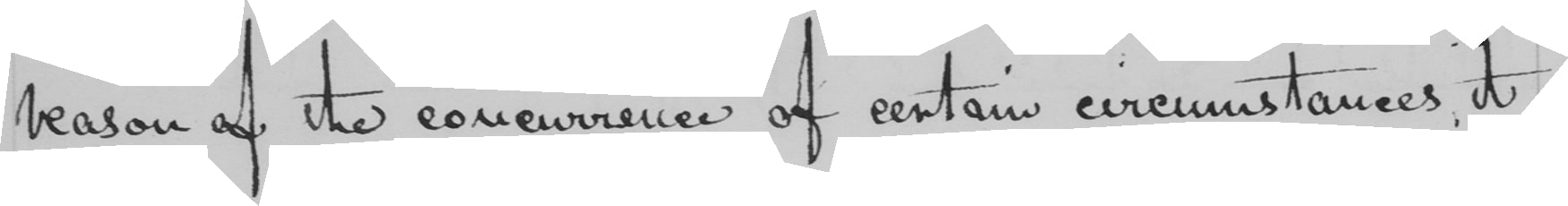}} & & \multicolumn{2}{c}{\includegraphics[width=0.45\linewidth]{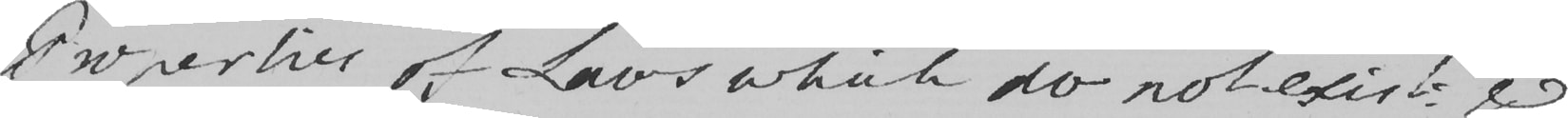}} \\
\textbf{Ground Truth}         & reason of the concurrence of certain circumstances, it & & 
\textbf{Ground Truth}         & Properties of Laws which do not exist \\
\cmidrule{1-2} \cmidrule{4-5}
\textbf{StdConv 1D-LSTM} & teason of the concnnence of antans ciraummstances to & & 
\textbf{StdConv 1D-LSTM} & Pwperlies of Laws which so nobeist: es \\
\textbf{DefConv 1D-LSTM}      & teason of the covernence of certais circumstances, to & & 
\textbf{DefConv 1D-LSTM}      & Parertics or Laws which do rot exish a \\
\textbf{StdConv CRNN}    & reason of the concnrences of antrvie ciroummnstances, to & & 
\textbf{StdConv CRNN}    & Pwrerties oI Laws which do notedist: \\
\textbf{DefConv CRNN}         & reson of the conerrence of cantain circumntances, th & & 
\textbf{DefConv CRNN}         & Pwperties of Laws which do notexist a \\

\addlinespace[0.24cm]
\multicolumn{5}{c}{\textbf{ICFHR16 dataset}} \\
\multicolumn{2}{c}{\includegraphics[width=0.45\linewidth]{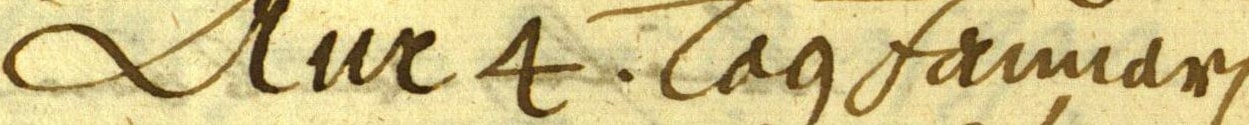}} & & \multicolumn{2}{c}{\includegraphics[width=0.45\linewidth]{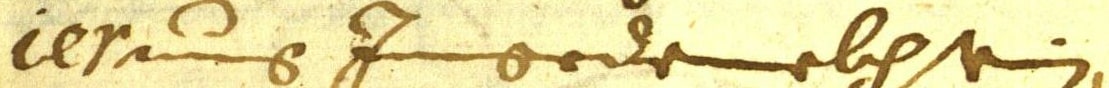}} \\
\textbf{Ground Truth}         & Am 4. Tag Janūarj \& & & 
\textbf{Ground Truth}         & ierūng Inngedennckh sein, \\
\cmidrule{1-2} \cmidrule{4-5}
\textbf{StdConv 1D-LSTM} & Am 4. Tag Janarj & & 
\textbf{StdConv 1D-LSTM} & ilsūng Inngedenckh sein., \\
\textbf{DefConv 1D-LSTM}      & Am4. Tag Kamarj & & 
\textbf{DefConv 1D-LSTM}      & ierūng Inngedennckh sein, \\
\textbf{StdConv CRNN}    & Am4. Tag Janmarj & & 
\textbf{StdConv CRNN}    & ilūnng Inngedenckh sein.  \\
\textbf{DefConv CRNN}         & Am: Tag Jamarj & & 
\textbf{DefConv CRNN}         & ierūng Inngedennckh sein. \\

\addlinespace[0.24cm]
\multicolumn{5}{c}{\textbf{Leopardi dataset}} \\
\multicolumn{2}{c}{\includegraphics[width=0.45\linewidth]{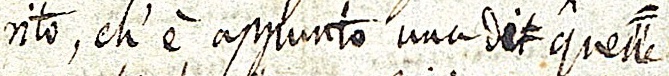}} & & \multicolumn{2}{c}{\includegraphics[width=0.45\linewidth]{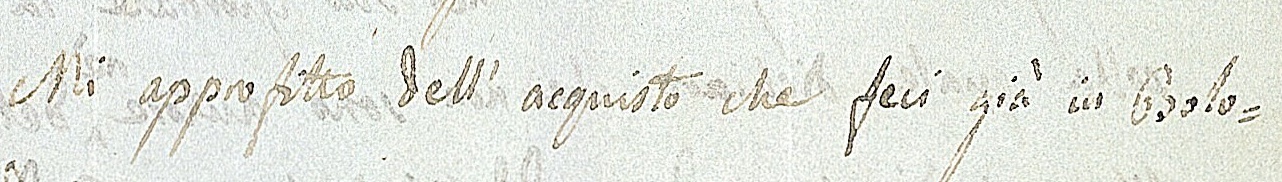}} \\
\textbf{Ground Truth}         & rito, ch'è appunto una di quelle & & 
\textbf{Ground Truth}         & Mi approfitto dell'acquisto che feci già in Bolo= \\
\cmidrule{1-2} \cmidrule{4-5}
\textbf{StdConv 1D-LSTM} & rito, ch' è apglunto una det queti= & & 
\textbf{StdConv 1D-LSTM} & Mi oppofilto dell'aspisto che fai già in biril= \\
\textbf{DefConv 1D-LSTM}      & rito, ch' è aggunto una dil quelle & & 
\textbf{DefConv 1D-LSTM}      & Mi pprofito dell'apisto che fei già in Boolo= \\
\textbf{StdConv CRNN}    & rito, ch' è apunto unla dil quelle & & 
\textbf{StdConv CRNN}    & Mi pogio dell'aluisto che fuai giù in dolo.= \\
\textbf{DefConv CRNN}         & rito, ch'è appunto una di quelle & & 
\textbf{DefConv CRNN}         & Mi gpro siteo dell'aguisto che feui giò in Bolo \\
\end{tabular}
}
\caption{Qualitative results on the historical datasets considered.}
\label{fig:hist_qualitatives}
\end{figure*}

With respect to the considered state-of-the-art approaches, on the IAM and RIMES datasets, the DefConv-based models perform competitively. 
Compared to the StandarConv-based baselines, the proposed models allow decreasing both the CER and the WER. Note that the improvement in terms of WER is more evident. This suggests that, by capturing more context, DefConv-based models make character-level errors that are more concentrated in fewer difficult words. The improvement is even more evident on the historical datasets. This indicates that our approach is more robust to background non-idealities, which can be observed also from the qualitative results reported in Figure~\ref{fig:modern_qualitatives} and Figure~\ref{fig:hist_qualitatives}.
This confirms that DefConvs are more robust to the nuisances present in the input image since their activations are more concentrated on the writing. 

The two proposed variants share the general HTR paradigm of convolutional-recurrent text image representation and CTC-based decoding of the transcription. The main difference between the two is in the number of convolutional and recurrent layers. While 1D-LSTM has a deeper recurrent component, CRNN has a deeper convolutional component. The larger number of recurrent layers makes the 1D-LSTM comparable or better than the CRNN variant when both their convolutional components contain standard convolutions. On the other hand, due to the larger number of convolutional layers in CRNN, this variant outperforms the other when both are equipped with DefConvs. In general, the CRNN variant benefits the most from the introduction of DefConvs. In fact, on average, its CER decreases by $0.7$ and the WER by $2.5$, while for the 1D-LSTM variant, the CER decreases by $0.6$ and the WER by $1.0$. This is more evident on the historical datasets, where the background is noisier due to the aging of the page support. From this observation, we can conclude that a larger number of DefConv layers improves the robustness to background noise.

The results of the pretraining plus fine-tuning experiments on the Leopardi dataset are reported in Table~\ref{tab:fine-tuning}. 
It emerges that DefConvs allow better exploiting this paradigm, as confirmed by the comparable to smaller errors obtained with the variants featuring DefConvs in all settings. Moreover, the performance gap between pre-training on historical synthetic data and on modern synthetic data is smaller when using DefConv-based models. This suggests that models based on DefConvs can generalize better than their StdConv-based counterparts and would be more effective for HTR of small historical manuscript collections, even if pretrained on general, modern-looking data.


\section{Conclusion}
In this work, we investigated the suitability of deformable convolutions for the HTR task. We validated our approach on both modern and historical datasets of various languages and historical periods and demonstrated their superior performance with respect to standard convolutions.
Due to the ability to adapt to highly distorted handwritten strokes and focus on ink pixels while still capturing more context, DefConv-based HTR models are effective when dealing with free-layout documents and allow better exploiting the pretraining plus fine-tuning paradigm for HTR of small collections of historical documents.
Further performance improvements could be achieved by employing a language-specific language model, a direction which we leave for future investigation.


\section*{Declarations}
This work was supported by the ``AI for Digital Humanities'' project (Pratica Sime n.2018.0390), funded by ``Fondazione di Modena'', and by the ``DHMoRe Lab'' project (CUP E94I19001060003), funded by ``Regione Emilia Romagna''.

\bibliographystyle{sn-chicago}
\bibliography{bibliography}


\begin{appendices}
\clearpage
\section*{Appendix}
\subsection*{Models Architectures}
We provide the detailed architecture of the proposed DefConv-based HTR models: in Table~\ref{tab:crnn} for CRNN, in Table~\ref{tab:1d-lstm} for 1D-LSTM. The offsets of the DefConvs layers are handled in a standard convolutional layer before the DefConv, which is in charge of learning two parameters for each kernel cell of the DefConv. Note that the output size of the final Linear layer, $c$ depends on the charset size of each dataset (including the \emph{blank} character). In particular: $c = 96$ for the IAM dataset, $c = 80$ for the RIMES,  $c = 89$ for the ICFHR14, $c = 94$ for the ICFHR16, and $c = 77$ for the Leopardi. Note that, from a practical standpoint, when the whole dataset is available, $c$ can be calculated directly as the number of the characters appearing in the dataset (\ie~the charset), plus the $blank$ character. For new or unknown datasets, the charset, and thus, $c$, can be estimated \eg~from large corpora in the same language as the dataset of interest, but can potentially include as many characters as the designer wants. In this latter case, logits corresponding to characters included in the charset but not appearing in the dataset of interest will be assigned zero probability. 
In the StandardConv-based baselines we used in the experiments, each pair of offset Convolution layer and the DefConv layer is replaced by a standard convolution layer with the same characteristics as the DefConv layer.

\begin{table}[tb]
\centering
\setlength{\tabcolsep}{0.35em}
\resizebox{0.95\columnwidth}{!}{%
\begin{tabular}{ccccc}
\toprule
\textbf{Layer Type} & \textbf{Size} & \textbf{Kernel} & \textbf{Stride} & \textbf{Padding} \\
\toprule
Convolution & 18 & 3 $\times$ 3 & (1, 1) & (1, 1) \\
DefConv & 64 & 3 $\times$ 3 & (1, 1) & (1, 1) \\
Batch Normalization & $-$ & $-$ & $-$ & $-$ \\
ReLU & $-$ & $-$ & $-$ & $-$ \\
Max Pooling & $-$ & 2 $\times$ 2 & (2, 2) & (0, 0) \\
Dropout (p = 0.2) & $-$ & $-$ & $-$ & $-$ \\
\midrule
Convolution & 18 & 3 $\times$ 3 & (1, 1) & (1, 1) \\
DefConv & 128 & 3 $\times$ 3 & (1, 1) & (1, 1) \\
Batch Normalization & $-$ & $-$ & $-$ & $-$ \\
ReLU & $-$ & $-$ & $-$ & $-$ \\
Max Pooling & $-$ & 2 $\times$ 2 & (2, 2) & (0, 0) \\
Dropout (p = 0.2) & $-$ & $-$ & $-$ & $-$ \\
\midrule
Convolution & 18 & 3 $\times$ 3 & (1, 1) & (1, 1) \\
DefConv & 256 & 3 $\times$ 3 & (1, 1) & (1, 1) \\
Batch Normalization & $-$ & $-$ & $-$ & $-$ \\
ReLU & $-$ & $-$ & $-$ & $-$ \\
\midrule
Convolution & 18 & 3 $\times$ 3 & (1, 1) & (1, 1) \\
DefConv & 256 & 3 $\times$ 3 & (1, 1) & (1, 1) \\
ReLU & $-$ & $-$ & $-$ & $-$ \\
Max Pooling & $-$ & 2 $\times$ 2 & (2, 1) & (0, 1) \\
Dropout (p = 0.2) & $-$ & $-$ & $-$ & $-$ \\
\midrule
Convolution & 18 & 3 $\times$ 3 & (1, 1) & (1, 1) \\
DefConv & 512 & 3 $\times$ 3 & (1, 1) & (1, 1) \\
Batch Normalization & $-$ & $-$ & $-$ & $-$ \\
ReLU & $-$ & $-$ & $-$ & $-$ \\
Dropout (p = 0.2) & $-$ & $-$ & $-$ & $-$ \\
\midrule
Convolution & 18 & 3 $\times$ 3 & (1, 1) & (1, 1) \\
DefConv & 512 & 3 $\times$ 3 & (1, 1) & (1, 1) \\
ReLU & $-$ & $-$ & $-$ & $-$ \\
Max Pooling & $-$ & 2 $\times$ 2 & (2, 1) & (0, 1) \\
Dropout (p = 0.2) & $-$ & $-$ & $-$ & $-$ \\
\midrule
Convolution & 8 & 2 $\times$ 2 & (1, 1) & (0, 0) \\
DefConv & 512 & 2 $\times$ 2 & (1, 1) & (0, 0) \\
Batch Normalization & $-$ & $-$ & $-$ & $-$ \\
ReLU & $-$ & $-$ & $-$ & $-$ \\
\midrule
\midrule
BLSTM & 512 & $-$ & $-$ & $-$ \\
\midrule
Dropout (p = 0.5) & $-$ & $-$ & $-$ & $-$ \\
\midrule
BLSTM & 512 & $-$ & $-$ & $-$ \\
\midrule
Linear & $c$ & $-$ & $-$ & $-$ \\
\bottomrule
\end{tabular}%
}
\caption{Architecture details for the CRNN variant.}
\label{tab:crnn}
\end{table}

\begin{table}[tb]
\centering
\setlength{\tabcolsep}{0.35em}
\resizebox{0.95\columnwidth}{!}{%
\begin{tabular}{ccccc}
\toprule
\textbf{Layer Type} & \textbf{Size} & \textbf{Kernel} & \textbf{Stride} & \textbf{Padding} \\
\toprule
Convolution & 18 & 3 $\times$ 3 & (1, 1) & (1, 1) \\
DefConv & 16 & 3 $\times$ 3 & (1, 1) & (1, 1) \\
Batch Normalization & $-$ & $-$ & $-$ & $-$ \\
LeakyReLU & $-$ & $-$ & $-$ & $-$ \\
Max Pooling & $-$ & 2 $\times$ 2 & (2, 2) & (0, 0) \\
\midrule
Convolution & 18 & 3 $\times$ 3 & (1, 1) & (1, 1) \\
DefConv & 32 & 3 $\times$ 3 & (1, 1) & (1, 1) \\
Batch Normalization & $-$ & $-$ & $-$ & $-$ \\
LeakyReLU & $-$ & $-$ & $-$ & $-$ \\
Max Pooling & $-$ & 2 $\times$ 2 & (2, 2) & (0, 0) \\
Dropout (p = 0.2) & $-$ & $-$ & $-$ & $-$ \\
\midrule
Convolution & 18 & 3 $\times$ 3 & (1, 1) & (1, 1) \\
DefConv & 48 & 3 $\times$ 3 & (1, 1) & (1, 1) \\
Batch Normalization & $-$ & $-$ & $-$ & $-$ \\
LeakyReLU & $-$ & $-$ & $-$ & $-$ \\
Max Pooling & $-$ & 2 $\times$ 2 & (2, 2) & (0, 0) \\
Dropout (p = 0.2) & $-$ & $-$ & $-$ & $-$ \\
\midrule
Convolution & 18 & 3 $\times$ 3 & (1, 1) & (1, 1) \\
DefConv & 64 & 3 $\times$ 3 & (1, 1) & (1, 1) \\
Batch Normalization & $-$ & $-$ & $-$ & $-$ \\
LeakyReLU & $-$ & $-$ & $-$ & $-$ \\
Dropout (p = 0.2) & $-$ & $-$ & $-$ & $-$ \\
\midrule
Convolution & 18 & 3 $\times$ 3 & (1, 1) & (1, 1) \\
DefConv & 80 & 3 $\times$ 3 & (1, 1) & (1, 1) \\
Batch Normalization & $-$ & $-$ & $-$ & $-$ \\
LeakyReLU & $-$ & $-$ & $-$ & $-$ \\
\midrule
\midrule
BLSTM & 256 & $-$ & $-$ & $-$ \\
\midrule
Dropout (p = 0.5) & $-$ & $-$ & $-$ & $-$ \\
\midrule
BLSTM & 256 & $-$ & $-$ & $-$ \\
\midrule
Dropout (p = 0.5) & $-$ & $-$ & $-$ & $-$ \\
\midrule
BLSTM & 256 & $-$ & $-$ & $-$ \\
\midrule
Dropout (p = 0.5) & $-$ & $-$ & $-$ & $-$ \\
\midrule
BLSTM & 256 & $-$ & $-$ & $-$ \\
\midrule
Dropout (p = 0.5) & $-$ & $-$ & $-$ & $-$ \\
\midrule
BLSTM & 256 & $-$ & $-$ & $-$ \\
\midrule
Dropout (p = 0.5) & $-$ & $-$ & $-$ & $-$ \\
\midrule
Linear & $c$ & $-$ & $-$ & $-$ \\
\bottomrule
\end{tabular}%
}
\caption{Architecture details for the 1D-LSTM variant.}
\label{tab:1d-lstm}
\end{table}
\end{appendices}

\end{document}